\documentclass[10pt]{article}
\usepackage[letterpaper, margin=1in, top=1in, bottom=1in]{geometry}
\usepackage{times}

\usepackage{titlesec}
\titlespacing*{\section}{0pt}{1.2ex plus 0.3ex minus 0.2ex}{0.6ex plus 0.2ex}
\titlespacing*{\subsection}{0pt}{0.9ex plus 0.2ex minus 0.1ex}{0.4ex plus 0.1ex}
\titleformat{\section}[hang]{\normalfont\large\bfseries}{\thesection}{0.6em}{}
\titleformat{\subsection}[hang]{\normalfont\normalsize\bfseries}{\thesubsection}{0.5em}{}
\usepackage[skip=4pt, font=small, labelfont=bf]{caption}
\usepackage[utf8]{inputenc}
\usepackage[T1]{fontenc}
\usepackage{url}
\usepackage{booktabs}
\usepackage{array}
\usepackage{longtable}
\usepackage{calc}
\usepackage{etoolbox}
\usepackage{amsfonts}
\usepackage{amsmath}
\usepackage{amssymb}
\usepackage{amsthm}
\usepackage{nicefrac}
\usepackage{microtype}
\usepackage[round]{natbib}
\usepackage{xcolor}
\usepackage{graphicx}
\usepackage[hidelinks]{hyperref}
\graphicspath{{figures/}}

\usepackage{newunicodechar}
\newunicodechar{×}{\ensuremath{\times}}
\newunicodechar{†}{\ensuremath{\dagger}}
\newunicodechar{≈}{\ensuremath{\approx}}
\newunicodechar{≥}{\ensuremath{\geq}}
\newunicodechar{§}{\S}
\newunicodechar{→}{\ensuremath{\rightarrow}}
\providecommand{\LTcaptype}{table}

\providecommand{\tightlist}{\setlength{\itemsep}{0pt}\setlength{\parskip}{0pt}}

\title{CIVeX: Causal Intervention Verification for Language Agents}
\author{Fabio Rovai, The Tesseract Academy \\ \texttt{fabio@thetesseractacademy.com}}

\begin{document}
\maketitle

\begin{abstract}
A valid tool call is not necessarily a valid intervention. Tool-using
language agents are guarded by schema validators, policy filters,
provenance checks, state predictors, and self-verification, yet such
safeguards do not certify that a state-changing action has an
identifiable causal effect. In confounded workflows, the action that
looks optimal in observational logs can reduce utility when executed.

We introduce CIVeX, a novel causal intervention verifier that maps
proposed actions to structural causal queries over a committed
action-state graph, checks identifiability, and returns one of four
auditable verdicts: EXECUTE, REJECT, EXPERIMENT, or ABSTAIN. Execution
requires an assumption-scoped causal certificate carrying graph
commitments, an identification argument, a one-sided lower confidence
bound (LCB), provenance, and risk limits.

On Causal-ToolBench (1,890 instances, 7 seeds), CIVeX yields zero
observed false executions across moderate and adversarial confounding.
Under adversarial confounding it reaches 84.9\% accuracy and 81.1\% of
oracle utility (+2.23 vs +2.76; 95\% CI {[}2.16, 2.31{]}) and is the
only non-oracle method whose constrained utility under a hard
zero-false-execution constraint exceeds the AlwaysAbstain floor of
+0.99. On IHDP (semi-synthetic) and ZOZO Open Bandit (real production
logs with uniform-random ground truth), CIVeX matches Oracle
correct-execution within 0.1pp and cuts per-execute false-execution by
≥50× over naive baselines. A chain-of-thought LLM verifier (Claude Opus
and Sonnet) cuts false-execution by an order of magnitude over a terse
baseline, yet under adversarial confounding Opus's utility falls to 74\%
of CIVeX's and Sonnet retains 1.0\% false-execution.

Two propositions characterise the per-instance error: any verifier
deciding from observational sign incurs false-execution rate at least
the trap fraction in confounded environments; CIVeX's per-execute
false-execution probability is bounded by the one-sided LCB level when
the committed graph is correct. Intervention identifiability, not action
validity, is the missing primitive for reliable tool use, conditional on
a sound graph-commitment infrastructure.
\end{abstract}

\section{Introduction}

Tool-using language agents now drive systems that delete data, execute
trades, deploy code, and modify databases. The standard safety stack for
such agents is built from validation primitives: schema validators that
check whether a tool call is well-formed \citep{yao2023react}, policy
filters that check whether the call is allowed, provenance trackers that
record where inputs came from, state predictors that forecast the
post-call state, and self-verification prompts
\citep{shinn2023reflexion, madaan2023self} that ask the agent to
critique its own action. Each of these primitives answers a different
question, but none answers the question that matters once the action
runs in a confounded world: does taking this action change the outcome
in the way the agent expects?

A tool call is not the same as an intervention
\citep{pearl2009causality}. The schema can be valid, the policy can
permit the action, the provenance can be clean, and the predicted next
state can be plausible, while the action's true causal effect on the
outcome of interest is biased, unidentifiable, or harmful. In
observational logs, an action that correlates with high utility may
produce low utility when actually performed, because the historical
correlation was driven by latent factors that an autonomous decision
system has no reason to know about. This is the textbook problem of
confounding \citep{hernan2020causal}, and it is the failure mode that
current agent safety stacks do not address.

We propose to treat tool execution as causal inference. Before a
state-changing action fires, we require an assumption-scoped causal
certificate: an explicit commitment to a graph over the action and state
variables, an identification or bounding argument that licenses an
estimate of the action's effect on the utility variable, provenance
metadata for the data used in that estimate, an uncertainty constraint
expressed as a one-sided lower confidence bound, and a risk limit on
potential harm. We call this object a \emph{causal certificate} and the
verifier that issues or refuses certificates \emph{CIVeX} (Causal
Intervention Verifier).

CIVeX maps a proposed action to a structural causal query of the form
\(E[Y \mid \mathrm{do}(T = t)]\), evaluates whether this query is
identifiable under the committed graph using a finite set of standard
tools (backdoor adjustment, frontdoor adjustment, or instrumental
variables when applicable), and routes the decision into one of four
auditable buckets: EXECUTE, REJECT, EXPERIMENT, or ABSTAIN. Each verdict
carries a certificate (or a refusal explaining the missing assumption),
so an auditor can replay the decision after the fact.

This is a deliberately narrow contract. CIVeX does not predict actions;
it gates them. It does not learn graphs from data; it commits to a graph
and checks identifiability. It does not replace existing validators; it
adds the layer they do not provide. CIVeX does not, on its own,
guarantee safety: the safety guarantee is conditional on the committed
graph being correct, an assumption that in production must be supported
by a graph-commitment infrastructure (versioning, signing, drift
monitoring) outside the scope of this paper.

We make three contributions. First, we formalise the causal certificate
object and the four-way decision in Section \hyperref[3-civex]{3}, and
give a triage algorithm whose outputs are auditable. Second, we provide
two propositions in Section \hyperref[4-theoretical-evidence]{4}
characterising when non-causal verification is unsafe and bounding
CIVeX's false-execution rate; the propositions are statements about
per-instance error rates, not asymptotic regret. Third, we introduce
Causal-ToolBench in Section \hyperref[5-causal-toolbench]{5}, a
benchmark of six tool-using workflow families instantiated as structural
causal models (SCMs) with explicit confounding strength, counterbalanced
safety and harm, paired randomised controlled trial (RCT)--style data
for stage-2 evaluation, and an adversarial regime in which the hidden
confounder is calibrated to flip the sign of the observational
association. Section \hyperref[6-experiments]{6} reports the empirical
evaluation, including four sensitivity sweeps and a chain-of-thought
LLM-baseline comparison.

\section{Related work}

\textbf{Causal identification.} The do-calculus
\citep{pearl2009causality} and the Shpitser-Pearl ID algorithm
\citep{shpitser2006identification} determine whether a query of the form
\(E[Y \mid \mathrm{do}(T)]\) is identifiable from observational data
given a graph. Backdoor and frontdoor adjustment
\citep{tian2002general, pearl2009causality} are the most common
identification arguments; partial identification via Manski bounds
\citep{manski2003partial} and instrumental variables provide the
alternatives we discuss in Section \hyperref[33-triage]{3.3}.
Sensitivity analysis
\citep{rosenbaum1983central, vanderweele2017sensitivity, cinelli2020making}
quantifies how badly an unmeasured confounder would have to behave to
overturn an identified effect. The certificate object generalises these
tools into a single auditable artefact for runtime use.

\textbf{Off-policy evaluation under confounding.} Policy evaluation with
hidden confounders is studied in
\citep{kallus2018confounding, namkoong2020off, bareinboim2015bandits, tennenholtz2020off};
the doubly-robust / debiased-ML literature
\citep{chernozhukov2018double, kennedy2022semiparametric} gives
consistent estimators under partial confounding. The
``z-identifiability'' framework \citep{bareinboim2012causal} formalises
which target queries are identifiable from the union of observational
and experimental distributions, which is the exact setting CIVeX's
EXPERIMENT branch invokes. CIVeX is not a policy-evaluation method; it
is a per-action runtime gate that consumes the same identification
machinery and refuses to issue a certificate when identifiability fails.

\textbf{Tool-using language agents.} ReAct \citep{yao2023react},
Reflexion \citep{shinn2023reflexion}, Self-Refine
\citep{madaan2023self}, and Toolformer \citep{schick2023toolformer}
propose mechanisms by which agents reason about and use tools.
Self-consistency \citep{wang2023self} and chain-of-thought prompting
\citep{wei2022chain} are widely used to improve answer quality. We treat
these methods as orthogonal to CIVeX: they govern internal coherence of
the agent's reasoning, not whether the resulting intervention has an
identifiable causal effect.

\textbf{Causal reasoning in LLMs.} Several benchmarks and analyses
\citep{kiciman2023causal, jin2023cladder, zecevic2023causal} evaluate
whether LLMs perform causal reasoning. Our LLM baselines in Section
\hyperref[65-llm-baselines-engagement-without-safety]{6.5} test whether
a strong model, given a graph and a chain-of-thought prompt with
two-shot exemplars, can be relied upon as an identification verifier.
The empirical answer is that it engages but is not safe.

\textbf{Safe RL and constrained decision-making.} Constrained MDPs
\citep{altman1999constrained, achiam2017constrained}, shielded RL
\citep{alshiekh2018safe}, and safe-exploration surveys
\citep{garcia2015comprehensive} provide algorithms that respect
constraints during execution. CIVeX shares the spirit of ``act only when
justified'' but operates at the granularity of a single tool call from
an off-the-shelf agent rather than a learned policy, and the certificate
is auditable rather than embedded in a learned controller.

\textbf{Audit and accountability.} Algorithmic-auditing frameworks
\citep{raji2020closing} and model cards \citep{mitchell2019model}
motivate per-action explainability. The certificate object is a
per-decision audit artefact that documents the assumptions under which
an action was permitted.

\section{CIVeX}

\subsection{Action frames and causal
queries}

A proposed action is represented as an \emph{action frame}
\(a = (\mathrm{tool}, t_v, t^*, Y, c, r)\), where \(\mathrm{tool}\) is
the tool name, \(t_v\) is the \emph{target variable} on which the action
intervenes, \(t^*\) is the value to which the agent intends to set the
target, \(Y\) is the \emph{utility variable}, \(c\) is the action's
expected cost (in the same units as utility), and \(r \in \{0, 1\}\)
indicates whether the action is reversible. We reserve
\(T \in \{0, 1\}\) for the binary treatment indicator (the on/off status
of \(t_v\)) and \(H\) for the planning horizon (number of CIVeX
invocations).

CIVeX maps \(a\) to the structural causal query
\(E[Y \mid \mathrm{do}(t_v = t^*)]\) and asks whether it is identifiable
from observational data under a committed graph \(G\).

\subsection{Causal certificate}

A causal certificate is a tuple
\(\mathcal{C} = (G, \mathcal{A}, \pi, \hat{\theta}, \mathrm{LCB}_\alpha, \mathrm{prov}, \mathrm{risk})\):

\begin{itemize}
\tightlist
\item
  \(G\) is the committed causal graph over
  \(\{T, Y\} \cup \mathrm{Cov}\), possibly including bidirected edges
  representing latent confounding.
\item
  \(\mathcal{A}\) is a labelled set of assumptions (Section
  \hyperref[4-theoretical-evidence]{4}).
\item
  \(\pi\) is a machine-checkable identification proof: a backdoor
  adjustment set, a frontdoor decomposition, an instrumental-variable
  argument, or a sensitivity bound.
\item
  \(\hat{\theta}\) is the point estimate of the do-effect.
\item
  \(\mathrm{LCB}_\alpha\) is a \emph{one-sided} \(1-\alpha\) lower
  confidence bound on \(\hat{\theta}\).
\item
  \(\mathrm{prov}\) is the provenance hash of the data used to form
  \(\hat{\theta}\) (we use SHA-256 of the row-ordered data frame).
\item
  \(\mathrm{risk}\) is the upper bound on potential harm under
  \(\mathcal{A}\).
\end{itemize}

CIVeX issues a certificate only if (i) the query is identifiable under
\(G\) via \(\pi\), (ii) \(\mathrm{LCB}_\alpha \geq \tau_u\) (utility
threshold), and (iii) \(c \leq \tau_r\) (risk threshold). The choice of
one-sided rather than two-sided is intentional and load-bearing: a
one-sided bound at level \(\alpha\) controls the probability of false
execution at \(\alpha\) without inflating the rate by a factor of two,
as we detail in Section \hyperref[4-theoretical-evidence]{4}.

\subsection{Triage}

Given an action frame \(a\), a graph set
\(\mathcal{G} = \{G_1, \ldots, G_K\}\) representing the agent's prior
over the true graph, and observational data \(D\), CIVeX returns one of
four decisions through the following rules.

\textbf{Rule 1 (Tool gate).} If \(a\) is non-interventional (an
observation or a pure computation), execute. If \(a\) is malformed or
the action's risk class is forbidden, reject.

\textbf{Rule 2 (Identification).} For each \(G_k \in \mathcal{G}\),
attempt to identify \(E[Y \mid \mathrm{do}(T = t^*)]\) via a backdoor or
frontdoor adjustment. (Other identification arguments are admissible; we
limit our implementation to backdoor and frontdoor for clarity.)
Partition \(\mathcal{G}\) into identified graphs
\(\mathcal{G}_{\mathrm{id}}\) and not-identified graphs
\(\mathcal{G}_{\mathrm{ni}}\).

\textbf{Rule 3 (Certified execution).} If
\(\mathcal{G}_{\mathrm{ni}} = \emptyset\), estimate \(\hat{\theta}\) and
a one-sided \(1-\alpha\) lower confidence bound \(\mathrm{LCB}_\alpha\)
from \(D\) using each adjustment set in \(\mathcal{G}_{\mathrm{id}}\),
and take the minimum (worst-case) LCB across graphs. If
\(\min_k \mathrm{LCB}^{(k)}_\alpha \geq \tau_u\) and \(c \leq \tau_r\),
issue a certificate and return EXECUTE. If
\(\min_k \mathrm{LCB}^{(k)}_\alpha < \tau_u\), return REJECT. If the LCB
clears \(\tau_u\) but \(c > \tau_r\), return REJECT (cost overruns the
risk threshold).

\textbf{Rule 4 (Bounded-risk experimentation).} If
\(\mathcal{G}_{\mathrm{id}} = \emptyset\), return EXPERIMENT when
\(c \leq \tau_r\) and \(r = 1\); ABSTAIN otherwise. Mixed sets route as
in Rule 4 (the unidentified branch dominates). We restrict to \(K = 1\)
in the evaluation, so the partition is always clean.

When EXPERIMENT is returned, the agent is expected to gather the
corresponding evidence and call CIVeX again with updated \(D\). In our
evaluation we simulate experimental evidence by re-running the verifier
on a paired dataset \(D^{\mathrm{exp}}\) that the SCM produces with
treatment randomly assigned (an idealised RCT). Section
\hyperref[64-ablation-experiment-branch-is-doing-real-work]{6.4} shows
that the certificate machinery and the EXPERIMENT branch each contribute
additively to CIVeX's utility advantage.

\subsection{Worked example}

Consider an \texttt{add\_index} action on a database table, with utility
variable \texttt{latency\_savings\_ms} (positive is better). The
committed graph \(G\) has directed edges from observed confounders
\texttt{query\_volume} and \texttt{write\_volume} to both the treatment
and the outcome, no bidirected edges, and a directed edge from the
treatment to the outcome. CIVeX computes:

\begin{enumerate}
\def\labelenumi{\arabic{enumi}.}
\tightlist
\item
  \textbf{Identification (Rule 2).} Backdoor adjustment on
  \(\{\text{query\_volume}, \text{write\_volume}\}\) identifies
  \(E[Y \mid \mathrm{do}(T=1)]\). \(\mathcal{G}_{\mathrm{id}} = \{G\}\),
  \(\mathcal{G}_{\mathrm{ni}} = \emptyset\).
\item
  \textbf{Estimation (Rule 3).} Ordinary least squares (OLS) regression
  of \texttt{latency\_savings\_ms} on \texttt{add\_index} plus the
  adjustment set yields \(\hat{\theta} = +3.10\) with one-sided 95\% LCB
  \(= +2.85\).
\item
  \textbf{Threshold check.}
  \(\mathrm{LCB}_{0.05} = +2.85 \geq \tau_u = 0\) and
  \(c = 0.05 \leq \tau_r = 0.5\).
\item
  \textbf{Certificate issued.}
  \(\mathcal{C} = (G, \mathcal{A}, \pi=\text{backdoor on \{query\_volume, write\_volume\}}, \hat{\theta}=3.10, \mathrm{LCB}_{0.05}=2.85, \mathrm{prov}=\text{SHA-256}, \mathrm{risk}=0.05)\).
  \textbf{Verdict: EXECUTE.}
\end{enumerate}

If the same instance had a bidirected edge \(T \leftrightarrow Y\) in
the graph (representing latent confounding), Rule 2 would put \(G\) in
\(\mathcal{G}_{\mathrm{ni}}\), identification would fail, and Rule 4
would route to EXPERIMENT (if reversible and within the cost budget) or
ABSTAIN. After collecting the experimental evidence, CIVeX re-runs on
the resolved graph and the RCT-style data; Rule 3 then recovers the
unbiased effect and issues the certificate.

\section{Theoretical evidence}

We give two propositions characterising the per-instance error behaviour
of observational and certified verifiers respectively. Full proofs are
in Appendix \hyperref[a-proofs-of-section-4]{A}. The propositions are
statements about per-instance false-execution rate, not asymptotic
regret in \(H\). We do not claim a sublinear-in-\(H\) regret bound; the
load-bearing claim is the bounded false-execution rate per execute.

\subsection{Assumptions}

We use a labelled assumption set throughout.

\begin{itemize}
\tightlist
\item
  \textbf{(A1) SUTVA / no interference.} The treatment of one instance
  does not affect the outcome of another.
\item
  \textbf{(A2) Graph correctness.} The committed graph \(G\) is a
  correct DAG of the underlying SCM.
\item
  \textbf{(A3) Positivity / overlap.}
  \(0 < P(T = t \mid \mathrm{Adj}) < 1\) for the adjustment set
  \(\mathrm{Adj}\) and treatment values of interest.
\item
  \textbf{(A4) Estimator consistency and CI validity.} The estimator
  \(\hat{\theta}\) is consistent for
  \(\theta = E[Y \mid \mathrm{do}(T=1)] - E[Y \mid \mathrm{do}(T=0)]\),
  and the one-sided \(1-\alpha\) lower confidence bound
  \(\mathrm{LCB}_\alpha\) has at least nominal coverage on the data
  \(D\) (under finite samples we use the OLS Wald-type interval).
\item
  \textbf{(A5) IID instances.} The benchmark instances are independent
  draws from the SCM family.
\end{itemize}

Proposition 1 uses (A1) and (A5). Proposition 2 uses (A1), (A2), (A3),
(A4).

\subsection{Non-causal verification is unsafe in confounded
environments}

Let \(\Pi_{\mathrm{obs}}\) be a verifier that issues EXECUTE based on
the sign of the empirical observational difference
\(\hat{\Delta} = \hat{E}[Y \mid T = 1] - \hat{E}[Y \mid T = 0]\)
exceeding zero. Let
\(\theta = E[Y \mid \mathrm{do}(T = 1)] - E[Y \mid \mathrm{do}(T = 0)]\)
be the true causal effect, and let \(b = \hat{\Delta} - \theta\) be the
observational-causal bias.

\textbf{Proposition 1 (informal, in expectation).} \emph{Define the}
harmful-trap fraction \(p_h > 0\) \emph{as the proportion of instances
on which the action is harmful (\(\theta < 0\)) yet the observational
sign-flip places \(\hat{\Delta} > 0\). Then the expected per-instance
false-execution rate of \(\Pi_{\mathrm{obs}}\) is at least \(p_h\), and
the expected total regret over \(H\) instances is at least
\(p_h \cdot H \cdot E[|\theta_i| \mid \text{harmful trap}]\).}

The bound restricts to harmful instances because sign-flips on safe
instances yield missed opportunities, not false executions. The bound is
linear in \(H\). Our adversarial regime (Section
\hyperref[53-two-regimes]{5.3}) constructs SCMs in which \(p_h\) is
positive by design.

\subsection{CIVeX safety bound}

Let \(\Pi_{\mathrm{CIVeX}}\) be the verifier of Section
\hyperref[33-triage]{3.3} with utility threshold \(\tau_u = 0\) and a
\emph{one-sided} \(1-\alpha\) lower confidence bound on
\(\hat{\theta}\).

\textbf{Proposition 2.} \emph{Under (A1)-(A4),
\(P(\mathrm{EXECUTE} \cap \theta < 0) \leq \alpha\). With \(K\)
candidate graphs including the true graph \(G^\star\) (by A2), the
worst-case rule (EXECUTE only if every
\(\mathrm{LCB}^{(k)}_\alpha \geq 0\)) inherits the same \(\alpha\) bound
without union correction, since execution implies an intersection of
safety conditions. A Bonferroni \(\alpha/K\) correction is required only
for the disjunctive variant (EXECUTE if any graph's LCB is
non-negative), which we do not use.}

This is a bound on the joint probability of executing-while-harmful, not
the conditional false-discovery rate; the proof is immediate from (A4).

\section{Causal-ToolBench}

\subsection{Workflow families}

The six families are: \texttt{db\_index\_operation},
\texttt{service\_restart\_operation}, \texttt{migration\_operation},
\texttt{cache\_operation}, \texttt{log\_retention\_operation}, and
\texttt{git\_branch\_operation}. Each family has the same action name
across all its instances, but each instance has a different context that
determines whether the action is causally beneficial or harmful. This
counterbalancing prevents methods that decide from the action name alone
(NameOnlyClassifier, schema-based gates) from achieving high accuracy by
name pattern matching. Per-family realised counterbalance ratios (Table
\hyperref[table-4-realised-counterbalance]{4} in Appendix
\hyperref[b-realised-counterbalance]{B}) show that each family's
safe-to-unsafe split is within \([0.40, 0.60]\) at every seed.

\subsection{Structural causal
models}

Each instance is sampled from a structural causal model:

\[
\begin{aligned}
T &= \mathrm{Bernoulli}\!\left(\mathrm{sigm}\!\left(\textstyle\sum_i \gamma_i\,\widetilde{U}_i\right)\right),\\
Y &= \beta_0 + \theta\,T + \textstyle\sum_i \beta_i\,\widetilde{U}_i + \varepsilon,\quad \varepsilon \sim \mathcal{N}(0, s^2),
\end{aligned}
\]

where \(U_i\) are observed and possibly hidden confounders,
\(\widetilde{U}_i = (U_i - \mu_i)/\sigma_i\) is the standardised version
used in both equations to keep coefficients on a common scale,
\(\mathrm{sigm}\) is the logistic function (we reserve \(\sigma\) for
confounder standard deviations), \(\gamma_i\) are the SCM coefficients
on the treatment logit (we use \(\gamma\) to avoid collision with the
LCB level \(\alpha\)), and \(\theta\) is the planted causal effect.

For every instance the generator returns the observational data frame,
the planted \(\theta\), the list of observed confounders, the list of
hidden confounders, and a \emph{paired} RCT-style data frame
\(D^{\mathrm{exp}}\) in which \(T\) is randomly assigned, used for
stage-2 evaluation when EXPERIMENT is returned. The full coefficient
ranges per family are in Table
\hyperref[table-5-scm-coefficient-ranges]{5} of Appendix
\hyperref[b-realised-counterbalance]{B}. Appendix
\hyperref[c-scm-recovery-checks]{C} verifies that adjusted OLS recovers
\(\theta\) within tolerance under moderate confounding and is biased by
construction under adversarial confounding.

\subsection{Two regimes (and a
sweep)}

\textbf{Moderate confounding.} The hidden confounder, when present, has
effect strengths \(|\gamma_h|, |\beta_h| \leq 0.6\). Adjusting on
observed covariates approximately recovers \(\theta\).
Observational-sign methods are usually correct.

\textbf{Adversarial confounding.} The hidden confounder has strength
\(|\gamma_h| = |\beta_h| = 2.5\). On harmful instances both are positive
(so \(U_h\) is positively correlated with both \(T\) and \(Y\), masking
the true negative effect); on safe instances \(\gamma_h\) and
\(\beta_h\) have opposite signs (so the spurious negative association
masks the true positive effect). This is calibrated so the observational
association sign-flips relative to \(\theta\) on the constructed
instances: a \emph{definitional} adversary, not an empirical one.
Section \hyperref[63-adversarial-strength-sweep]{6.3} reports a sweep
over the adversarial strength
\(|\gamma_h| = |\beta_h| \in \{0.5, 1.0, 1.5, 2.0, 2.5, 3.0, 3.5, 4.0\}\),
showing PolicyGate degrades monotonically and CIVeX is invariant.

\subsection{Counterbalancing and label
leakage}

Each action name is assigned to roughly equal numbers of safe and unsafe
instances per family. We report realised counterbalance ratios in
Appendix \hyperref[b-realised-counterbalance]{B}. Across all seeds and
families, NameOnlyClassifier, FamilyMajorityClassifier, and SchemaGate
achieve at most chance-level constrained-utility on the counterbalanced
families: any method that beats them must use more than the action name.

\subsection{Evaluation protocol}

Each method receives, for each instance, the action frame, the
observational data frame, and the graph set. Crucially, \emph{all
methods receive the same input}: causal methods (CIVeX,
CausalNoExperiment, ContextOnlyNoCausal, ObservationalAssociation) use
the graph; non-causal baselines (PolicyGate, SchemaGate,
FamilyMajorityClassifier, SemanticOntologyGate, NameOnlyClassifier,
AlwaysAbstain) ignore the graph by design but \emph{do receive it}. The
graph is not the experimental advantage; the experimental advantage is
whether the method consults it.

Stage-1 decisions are recorded. When a stage-1 decision is EXPERIMENT
and the instance has \texttt{safe\_experiment\_available\ =\ True}, the
method is re-invoked on the resolved graph (latent edge dropped) and the
paired \(D^{\mathrm{exp}}\) data, simulating a successful RCT. The
terminal decision is the result of the second invocation.

\subsection{Scoring}

Utility is a \emph{specified scoring rule} over the SCM-given causal
effect \(\theta\), not an intrinsic quantity. The SCM provides the
per-instance signed effect \(\theta\); the scoring function we apply on
top is our choice. Writing \(|\theta|\) for the magnitude and using the
safe/harmful label to set the sign, the rule is:

\[
u = \begin{cases}
+|\theta| - c_a & \text{EXECUTE on safe} \\
-|\theta| - c_a & \text{EXECUTE on harmful} \\
-w_{\mathrm{miss}}\, |\theta| & \text{ABSTAIN/REJECT on safe (missed opportunity)} \\
+|\theta| & \text{ABSTAIN/REJECT on harmful (correct refusal)}
\end{cases}
\]

with \(w_{\mathrm{miss}} = 0.3\) and action cost \(c_a = 0.05\) (denoted
\(c_{\mathrm{exp}}\) in the supplementary code). Because the scoring
rule is specified rather than given, Section
\hyperref[66-utility-weight-sweep]{6.6} sweeps
\(w_{\mathrm{miss}} \in \{0.0, 0.1, 0.3, 0.5, 1.0\}\) and
\(c_{\mathrm{exp}} \in \{0.0, 0.05, 0.25, 1.0\}\), showing CIVeX
dominates AlwaysAbstain in 20 of 20 weight configurations and PolicyGate
in 20 of 20; the headline ranking does not depend on our specific
choices.

We additionally report \textbf{constrained utility}: maximum utility
subject to a hard zero-false-execution constraint per seed. Methods with
non-zero false-execution rate on any seed are disqualified.

\section{Experiments}

\subsection{Setup}

We evaluate 14 methods on 7 random seeds (42 to 48), with 25 moderate
instances per family per seed (1,050 moderate total) and 20 adversarial
instances per family per seed (840 adversarial total), for 1,890 total
instances. A 15th \texttt{CIVeX\_3way} row produces byte-identical
decisions to \texttt{CIVeX\_5way} on every instance and is reported as a
single CIVeX row. LLM baselines use \texttt{claude-opus-4-7} and
\texttt{claude-sonnet-4-6} \citep{anthropic2024claude} via the Anthropic
API; the headline LLM row is the chain-of-thought two-shot configuration
described in Section
\hyperref[65-llm-baselines-engagement-without-safety]{6.5}. All
confidence intervals are 95\% bootstrap percentile intervals over the 7
seed-level means (5 seed-level means for the CausalPromptingCoT
adversarial rows; see footnote in Table \hyperref[table-2]{2}) with
2,000 resamples \citep{efron1993introduction}; the seeds are pre-fixed.
The percentile bootstrap on small \(n\) is known to undercover
\citep{efron1993introduction}; we additionally report Wilcoxon
signed-rank exact \(p\)-values for pairwise comparisons in Appendix
\hyperref[d-pairwise-comparisons]{D}.

\textbf{Methods.} We compare CIVeX against an oracle and twelve
baselines covering causal-without-experiment, policy/heuristic, and LLM
verifiers. \emph{OracleSCM} uses the SCM-given \(\theta\) and is the
upper bound. \emph{CausalNoExperiment}, \emph{ContextOnlyNoCausal},
\emph{ObservationalAssociation}, and \emph{AlwaysAbstain} span the
causal-baseline spectrum (LCB-only, context-only, sign-only,
refuse-all). \emph{PolicyGate}, \emph{SchemaGate},
\emph{SemanticOntologyGate}, \emph{FamilyMajorityClassifier}, and
\emph{NameOnlyClassifier} are policy/heuristic gates that ignore the
graph. \emph{CausalPromptingTerse} is the original single-prompt LLM
verifier (\texttt{max\_tokens\ =\ 20}); \emph{CausalPromptingCoT-Opus /
-Sonnet} are the chain-of-thought variants
(\texttt{max\_tokens\ =\ 512}, two LCB-aware exemplars) detailed in
Section \hyperref[65-llm-baselines-engagement-without-safety]{6.5}. All
methods receive the same inputs (action frame, observational data,
committed graph); the experimental advantage of CIVeX is whether the
verifier consults the graph, not whether the graph is available.
Method-by-method definitions are in Appendix
\hyperref[j-method-definitions]{J}.

\subsection{Headline results}

Tables \hyperref[table-1]{1} and \hyperref[table-2]{2} report the
per-method aggregate metrics across 7 seeds with 95\% bootstrap CIs.
Figure \hyperref[figure-1]{1} plots each method's utility across the two
regimes.

\paragraph{Table 1}\label{table-1}

Moderate confounding (n = 1,050).

{\def\LTcaptype{none} % do not increment counter
\begin{longtable}[]{@{}
  >{\raggedright\arraybackslash}p{(\linewidth - 8\tabcolsep) * \real{0.2000}}
  >{\raggedright\arraybackslash}p{(\linewidth - 8\tabcolsep) * \real{0.2000}}
  >{\raggedright\arraybackslash}p{(\linewidth - 8\tabcolsep) * \real{0.2000}}
  >{\raggedright\arraybackslash}p{(\linewidth - 8\tabcolsep) * \real{0.2000}}
  >{\raggedright\arraybackslash}p{(\linewidth - 8\tabcolsep) * \real{0.2000}}@{}}
\toprule\noalign{}
\begin{minipage}[b]{\linewidth}\raggedright
Method
\end{minipage} & \begin{minipage}[b]{\linewidth}\raggedright
False exec
\end{minipage} & \begin{minipage}[b]{\linewidth}\raggedright
Correct exec
\end{minipage} & \begin{minipage}[b]{\linewidth}\raggedright
Accuracy
\end{minipage} & \begin{minipage}[b]{\linewidth}\raggedright
Utility (95\% CI)
\end{minipage} \\
\midrule\noalign{}
\endhead
\bottomrule\noalign{}
\endlastfoot
OracleSCM & 0.0\% & 49.7\% & 100.0\% & +2.90 {[}+2.85, +2.96{]} \\
PolicyGate & 0.0\% & 49.7\% & 100.0\% & +2.90 {[}+2.85, +2.96{]} \\
\textbf{CIVeX} & \textbf{0.0\%} & \textbf{29.6\%} & \textbf{79.9\%} &
\textbf{+2.29 {[}+2.18, +2.38{]}} \\
CausalPromptingCoT-Opus & 0.2\% & 29.6\% & 79.7\% & +2.28 {[}+2.16,
+2.38{]} \\
ContextOnlyNoCausal & 0.0\% & 23.3\% & 73.6\% & +2.21 {[}+2.10,
+2.32{]} \\
CausalPromptingCoT-Sonnet & 2.3\% & 28.2\% & 76.2\% & +2.08 {[}+1.96,
+2.20{]} \\
CausalNoExperiment & 0.0\% & 20.8\% & 71.0\% & +2.04 {[}+1.93,
+2.12{]} \\
CausalPromptingTerse & 5.7\% & 26.6\% & 71.1\% & +1.87 {[}+1.79,
+1.94{]} \\
AlwaysAbstain & 0.0\% & 0.0\% & 50.3\% & +1.43 {[}+1.33, +1.53{]} \\
ObservationalAssociation & 0.0\% & 0.0\% & 50.3\% & +1.43 {[}+1.33,
+1.53{]} \\
NameOnlyClassifier & 7.6\% & 9.0\% & 51.7\% & +1.40 {[}+1.29,
+1.52{]} \\
FamilyMajorityClassifier & 50.3\% & 49.7\% & 49.7\% & -0.64 {[}-0.76,
-0.51{]} \\
SchemaGate & 50.3\% & 49.7\% & 49.7\% & -0.64 {[}-0.76, -0.51{]} \\
SemanticOntologyGate & 50.3\% & 49.7\% & 49.7\% & -0.64 {[}-0.76,
-0.51{]} \\
\end{longtable}
}

\emph{ObservationalAssociation} collapses to \emph{AlwaysAbstain} on
this benchmark because our implementation pairs the sign check with an
LCB filter (Appendix \hyperref[j-method-definitions]{J}). The three
policy-style gates (FamilyMajorityClassifier, SchemaGate,
SemanticOntologyGate) reduce to a single ``execute everything not on a
forbidden list'' rule; we report all three rows because they instantiate
distinct policy concepts.

\paragraph{Table 2}\label{table-2}

Adversarial confounding (n = 840).

{\def\LTcaptype{none} % do not increment counter
\begin{longtable}[]{@{}
  >{\raggedright\arraybackslash}p{(\linewidth - 8\tabcolsep) * \real{0.2000}}
  >{\raggedright\arraybackslash}p{(\linewidth - 8\tabcolsep) * \real{0.2000}}
  >{\raggedright\arraybackslash}p{(\linewidth - 8\tabcolsep) * \real{0.2000}}
  >{\raggedright\arraybackslash}p{(\linewidth - 8\tabcolsep) * \real{0.2000}}
  >{\raggedright\arraybackslash}p{(\linewidth - 8\tabcolsep) * \real{0.2000}}@{}}
\toprule\noalign{}
\begin{minipage}[b]{\linewidth}\raggedright
Method
\end{minipage} & \begin{minipage}[b]{\linewidth}\raggedright
False exec
\end{minipage} & \begin{minipage}[b]{\linewidth}\raggedright
Correct exec
\end{minipage} & \begin{minipage}[b]{\linewidth}\raggedright
Accuracy
\end{minipage} & \begin{minipage}[b]{\linewidth}\raggedright
Utility (95\% CI)
\end{minipage} \\
\midrule\noalign{}
\endhead
\bottomrule\noalign{}
\endlastfoot
OracleSCM & 0.0\% & 54.5\% & 100.0\% & +2.76 {[}+2.73, +2.78{]} \\
\textbf{CIVeX} & \textbf{0.0\%} & \textbf{39.4\%} & \textbf{84.9\%} &
\textbf{+2.23 {[}+2.16, +2.31{]}} \\
CausalPromptingCoT-Sonnet † & 1.0\% & 28.0\% & 72.2\% & +1.82 {[}+1.72,
+1.93{]} \\
CausalPromptingCoT-Opus † & 0.0\% & 20.0\% & 65.2\% & +1.65 {[}+1.55,
+1.79{]} \\
CausalPromptingTerse & 10.2\% & 39.4\% & 74.6\% & +1.57 {[}+1.44,
+1.69{]} \\
AlwaysAbstain & 0.0\% & 0.0\% & 45.5\% & +0.99 {[}+0.92, +1.05{]} \\
CausalNoExperiment & 0.0\% & 0.0\% & 45.5\% & +0.99 {[}+0.92,
+1.05{]} \\
ContextOnlyNoCausal & 0.0\% & 0.0\% & 45.5\% & +0.99 {[}+0.92,
+1.05{]} \\
ObservationalAssociation & 0.0\% & 0.0\% & 45.5\% & +0.99 {[}+0.92,
+1.05{]} \\
NameOnlyClassifier & 8.8\% & 7.9\% & 44.5\% & +0.78 {[}+0.70,
+0.88{]} \\
FamilyMajorityClassifier & 45.5\% & 54.5\% & 54.5\% & -0.03 {[}-0.12,
+0.07{]} \\
SchemaGate & 45.5\% & 54.5\% & 54.5\% & -0.03 {[}-0.12, +0.07{]} \\
SemanticOntologyGate & 45.5\% & 54.5\% & 54.5\% & -0.03 {[}-0.12,
+0.07{]} \\
PolicyGate & 28.3\% & 6.9\% & 24.0\% & -0.27 {[}-0.38, -0.17{]} \\
\end{longtable}
}

† CausalPromptingCoT adversarial rows aggregate 5 seeds (42-46) due to
API budget exhaustion at seed 47; all other rows aggregate 7 seeds. See
§\hyperref[61-setup]{6.1} and Appendix
\hyperref[f-llm-baseline-detail]{F}.

\begin{figure}[!htb]
\centering
\includegraphics[width=0.50\linewidth]{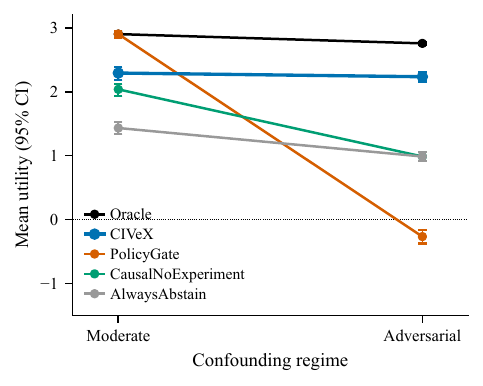}
\caption{Regime shift across two confounding regimes. PolicyGate ties OracleSCM under moderate confounding and collapses to negative utility under adversarial confounding. CIVeX is approximately invariant.}
\label{fig:1}
\end{figure}
\subsection{Adversarial-strength
sweep}

Figure \hyperref[figure-2]{2} shows how each method responds as the
hidden confounder strength \(|\gamma_h| = |\beta_h|\) varies (default
benchmark: \(2.5\)). CIVeX records zero observed false executions at
every strength tested; PolicyGate's false-execution rate increases
monotonically with confounder strength, from 0.0\% at \(|\gamma_h|=0.5\)
to 45.6\% at \(|\gamma_h|=4.0\). The full per-strength table is in
Appendix \hyperref[i-adversarial-strength-sweep-detail]{I}.

\begin{figure}[!htb]
\centering
\includegraphics[width=0.72\linewidth]{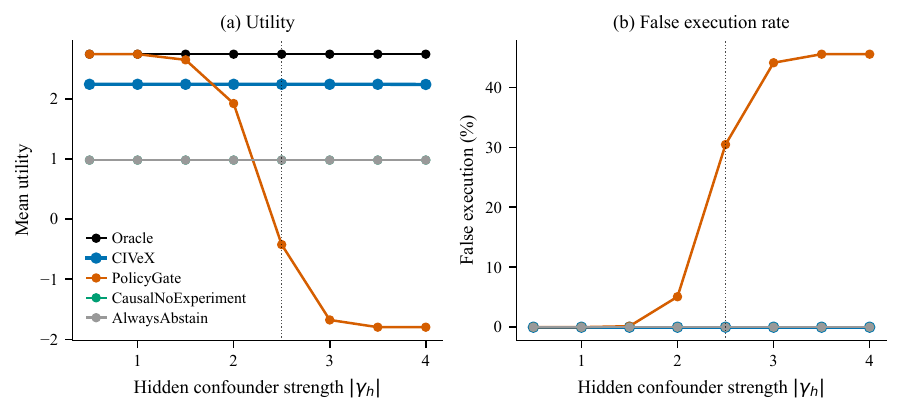}
\caption{Adversarial-strength sweep across \(|\gamma_h| \in \{0.5, 1.0, 1.5, 2.0, 2.5, 3.0, 3.5, 4.0\}\). Left panel: mean utility. Right panel: false-execution rate. CIVeX is approximately invariant; PolicyGate degrades monotonically. The vertical dotted line marks the default benchmark setting \(|\gamma_h| = 2.5\).}
\label{fig:2}
\end{figure}
\subsection{Ablation: EXPERIMENT branch is doing real
work}

An obvious concern is that the EXPERIMENT branch in our benchmark
consumes a paired RCT-style data frame \(D^{\mathrm{exp}}\) supplied by
the SCM. This is an idealised experiment, not a realistic exploration
cost model.

To isolate the certificate machinery from the experiment branch we
evaluate \textbf{CIVeX-CertOnly}: CIVeX with EXPERIMENT mapped to
ABSTAIN. CIVeX-CertOnly is byte-equivalent to the CausalNoExperiment
baseline. We report (Table \hyperref[table-6]{6} in Appendix
\hyperref[e-ablation-detail]{E}):

\begin{itemize}
\tightlist
\item
  \textbf{Moderate confounding}: CIVeX-CertOnly achieves +2.04 utility
  (zero false-exec), well above the AlwaysAbstain floor of +1.43.
  \emph{The certificate alone (LCB-aware EXECUTE/REJECT/ABSTAIN gating
  without any experimentation) contributes +0.61 utility.}
\item
  \textbf{Adversarial confounding}: CIVeX-CertOnly degenerates to
  AlwaysAbstain (+0.99). \emph{Under adversarial confounding the entire
  opportunity-capture gain of +1.25 utility comes from the EXPERIMENT
  branch, which depends on the availability of unbiased experimental
  data.}
\end{itemize}

\subsection{LLM baselines: engagement without
safety}

We compare two configurations. \textbf{CausalPromptingTerse} is a
single-prompt verifier with \texttt{max\_tokens\ =\ 20} that issues a
one-word verdict. \textbf{CausalPromptingCoT} uses chain-of-thought with
\texttt{max\_tokens\ =\ 512}, a system prompt establishing the safety
contract (including an explicit instruction to require a non-negative
LCB before EXECUTE), two worked-example exemplars from the moderate
regime, and tests both \texttt{claude-opus-4-7} and
\texttt{claude-sonnet-4-6} \citep{anthropic2024claude}. Full prompt and
exemplars in Appendix \hyperref[f-llm-baseline-detail]{F}. Both
configurations are supplied with the same committed graph and
observational summary that CIVeX uses; only the verifier differs.

\begin{figure}[!htb]
\centering
\includegraphics[width=0.70\linewidth]{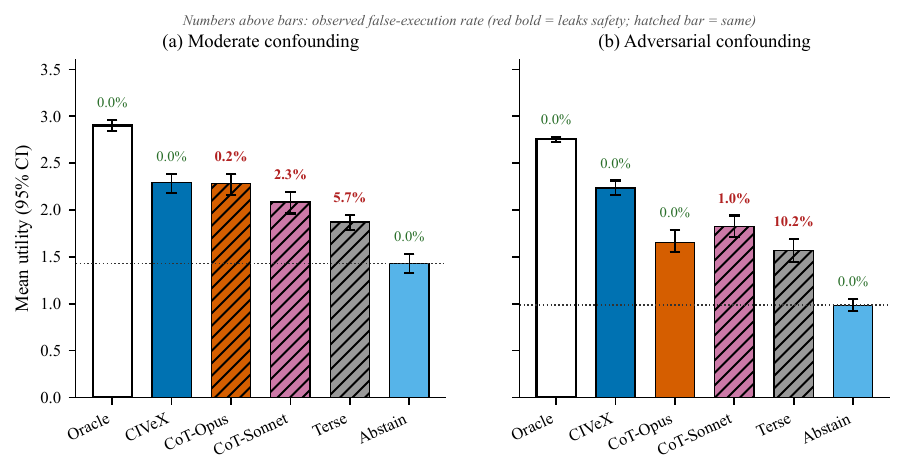}
\caption{LLM-improvement progression vs CIVeX in moderate (left) and adversarial (right) confounding. Bar height is mean utility with bootstrap 95\% CI; annotation above each bar is the observed false-execution rate (red bold = leaks safety; matching hatched bar). The Oracle bar is shown outlined-only because it is an upper bound, not a competing method. The dotted ``Abstain floor'' reference line is the AlwaysAbstain utility: methods below this line are worse than refusing every action. CoT-Opus matches CIVeX under moderate confounding but its adversarial utility falls to 74\% of CIVeX's; CoT-Sonnet retains a residual safety leak in both regimes.}
\label{fig:5}
\end{figure}
Numerical results are in Tables
\hyperref[table-1]{1}-\hyperref[table-2]{2} and broken down per
configuration in Table \hyperref[table-7]{7}. CoT-Opus matches CIVeX on
moderate confounding (+2.28 vs +2.29 utility, 0.2\% vs 0.0\% false-exec)
but its adversarial utility collapses to +1.65 (vs CIVeX's +2.23)
because it over-abstains on 35\% of instances CIVeX correctly
experiments through. CoT-Sonnet retains 2.3\% (moderate) and 1.0\%
(adversarial) false-exec, disqualified from the hard-safety
constrained-utility row of both tables. CoT cuts the terse baseline's
leak (5.7\% / 10.2\%) by roughly an order of magnitude on Opus but does
not eliminate it on Sonnet. The pattern is consistent with Section
\hyperref[4-theoretical-evidence]{4}: a deterministic identification
check enforces the LCB constraint by construction, while an LLM verifier
enforces it only when its reasoning step is correct. Strong prompting
raises the correct-reasoning rate but does not produce a per-instance
guarantee. The conclusion is deliberately narrow: on this benchmark,
with these two prompts and two models, an LLM verifier is not a
substitute for an algorithmic identification check under hard safety.

\subsection{Utility-weight sweep}

CIVeX dominates AlwaysAbstain in 20 of 20
\((w_{\mathrm{miss}}, c_{\mathrm{exp}})\) configurations spanning
\(w_{\mathrm{miss}} \in \{0.0, 0.1, 0.3, 0.5, 1.0\}\) and
\(c_{\mathrm{exp}} \in \{0.0, 0.05, 0.25, 1.0\}\), and dominates
PolicyGate in 20 of 20 under adversarial confounding. The headline
ranking does not depend on the chosen weights. Full table in Appendix
\hyperref[g-utility-weight-sweep]{G}.

\subsection{Graph misspecification}

We test CIVeX's robustness to graph misspecification by relabelling a
fraction of the observed confounders as latent (i.e., the variable
persists in the underlying SCM but is removed from the verifier's
committed graph and replaced by a bidirected \(T \leftrightarrow Y\)
edge representing the now-unblockable confounding path). Under 0\%,
25\%, 50\%, and 100\% relabel fractions (Appendix
\hyperref[h-graph-misspecification]{H}), CIVeX records zero observed
false executions; correct-execution drops from 32.4\% to 17.9\%.
Identification correctly fails on the affected instances and the
verifier routes them to EXPERIMENT or ABSTAIN.

\section{Discussion and
limitations}

We disclose six limitations. \textbf{(L1) Synthetic SCMs.}
Causal-ToolBench tests the \emph{property} that observational and causal
signs can disagree under hidden confounding; it does not match a
specific production system. \textbf{(L2) Graph commitment is the trusted
computing base.} CIVeX requires a correct committed graph; production
deployment needs a graph-commitment infrastructure (versioning, signing,
drift monitoring) outside this paper's scope. A plausible deployment
recipe would (i) elicit a candidate graph from domain experts (and where
available, prior RCT or natural-experiment evidence), (ii) validate
against held-out interventional data or assumption-bounding sensitivity
analyses, (iii) maintain a signed, timestamped graph registry with
semantic versioning, (iv) monitor for distributional drift on the
adjustment set, and (v) trigger re-validation when drift is flagged.
Each step is realisable with standard infrastructure
\citep{cinelli2020making}; this paper contributes the verifier, not the
pipeline. Section \hyperref[67-graph-misspecification]{6.7} shows CIVeX
retains zero observed false-exec under a single misspecification
mechanism. \textbf{(L3) Per-execute, not asymptotic regret.} Section
\hyperref[4-theoretical-evidence]{4} bounds per-instance error rates,
not regret in \(H\). The empirical false-execution rates (0/1,890 on
Causal-ToolBench and 10/7,470 = 0.13\% on IHDP) are both well below the
per-execute bound at \(\alpha = 0.05\); neither constitutes a
probability-of-zero claim outside the in-distribution evaluation.
\textbf{(L4) Small-\(n\) statistics.} The percentile bootstrap on seven
seed-level means is known to undercover \citep{efron1993introduction};
Appendix \hyperref[d-pairwise-comparisons]{D} reports Wilcoxon
signed-rank exact \(p\)-values. The rule-of-three
\citep{hanley1983if, eypasch1995probability} upper bound is 0.16\%
per-instance and \(3/7 = 42.9\%\) per-seed cluster; we recommend the
cluster-aware figure when generalising. \textbf{(L5) LLM
configuration-bounded.} The LLM finding is conditional on Claude Opus /
Sonnet with a two-shot LCB-aware CoT prompt; we do not test other model
families, paraphrases, or fine-tuned models. \textbf{(L6) Open
deployment threat model.} Graph injection, adjustment-set poisoning,
graph-version drift, and certificate revocation are real production
concerns we do not address; a serious deployment requires signed graphs
with provenance attestations, a graph registry, and a logging contract
between \(G\) and \(D\).

\section{Conclusion}

A tool call is not an intervention. CIVeX adds a four-way triage
(EXECUTE, REJECT, EXPERIMENT, ABSTAIN) that issues an auditable causal
certificate before execution, demands a bounded-risk experiment when the
effect is not identifiable, and abstains otherwise. Intervention
identifiability, not action validity, is the missing primitive.

Two extensions would close the gap to deployment. First, a
graph-commitment infrastructure realising the pipeline sketched in
Section \hyperref[7-discussion-and-limitations]{7} (L2):
expert-and-evidence elicitation, sensitivity-bounded validation, signed
registries, and drift monitoring. Second, a real-world case study on
logged production workflows with paired interventional data, replacing
the synthetic SCM regimes of Causal-ToolBench. Both build on the
verifier established here.

\bibliographystyle{plainnat}
\bibliography{references}

@book{pearl2009causality,
  title     = {Causality: Models, Reasoning, and Inference},
  author    = {Pearl, Judea},
  year      = {2009},
  edition   = {2nd},
  publisher = {Cambridge University Press},
  address   = {New York}
}

@book{hernan2020causal,
  title     = {Causal Inference: What If},
  author    = {Hern{\'a}n, Miguel A. and Robins, James M.},
  year      = {2020},
  address   = {Boca Raton},
  publisher = {Chapman \& Hall/CRC}
}

@inproceedings{shpitser2006identification,
  title     = {Identification of Joint Interventional Distributions in Recursive Semi-{M}arkovian Causal Models},
  author    = {Shpitser, Ilya and Pearl, Judea},
  booktitle = {Proceedings of the Twenty-First National Conference on Artificial Intelligence (AAAI)},
  pages     = {1219--1226},
  year      = {2006}
}

@inproceedings{bareinboim2012causal,
  title     = {Causal Inference by Surrogate Experiments: $z$-Identifiability},
  author    = {Bareinboim, Elias and Pearl, Judea},
  booktitle = {Proceedings of the 28th Conference on Uncertainty in Artificial Intelligence (UAI)},
  pages     = {113--120},
  year      = {2012}
}

@inproceedings{tian2002general,
  title     = {A General Identification Condition for Causal Effects},
  author    = {Tian, Jin and Pearl, Judea},
  booktitle = {Proceedings of the Eighteenth National Conference on Artificial Intelligence (AAAI)},
  pages     = {567--573},
  year      = {2002}
}

@article{rosenbaum1983central,
  title   = {The Central Role of the Propensity Score in Observational Studies for Causal Effects},
  author  = {Rosenbaum, Paul R. and Rubin, Donald B.},
  journal = {Biometrika},
  volume  = {70},
  number  = {1},
  pages   = {41--55},
  year    = {1983},
  doi     = {10.1093/biomet/70.1.41}
}

@article{vanderweele2017sensitivity,
  title   = {Sensitivity Analysis in Observational Research: Introducing the {E}-Value},
  author  = {VanderWeele, Tyler J. and Ding, Peng},
  journal = {Annals of Internal Medicine},
  volume  = {167},
  number  = {4},
  pages   = {268--274},
  year    = {2017},
  doi     = {10.7326/M16-2607}
}

@book{manski2003partial,
  title     = {Partial Identification of Probability Distributions},
  author    = {Manski, Charles F.},
  year      = {2003},
  publisher = {Springer},
  address   = {New York},
  series    = {Springer Series in Statistics}
}

@article{cinelli2020making,
  title   = {Making Sense of Sensitivity: Extending Omitted Variable Bias},
  author  = {Cinelli, Carlos and Hazlett, Chad},
  journal = {Journal of the Royal Statistical Society: Series B (Statistical Methodology)},
  volume  = {82},
  number  = {1},
  pages   = {39--67},
  year    = {2020},
  doi     = {10.1111/rssb.12348}
}

@inproceedings{bareinboim2015bandits,
  title     = {Bandits with Unobserved Confounders: A Causal Approach},
  author    = {Bareinboim, Elias and Forney, Andrew and Pearl, Judea},
  booktitle = {Advances in Neural Information Processing Systems 28 (NIPS)},
  year      = {2015}
}

@inproceedings{kallus2018confounding,
  title     = {Confounding-Robust Policy Improvement},
  author    = {Kallus, Nathan and Zhou, Angela},
  booktitle = {Advances in Neural Information Processing Systems 31 (NeurIPS)},
  year      = {2018}
}

@inproceedings{namkoong2020off,
  title     = {Off-Policy Policy Evaluation for Sequential Decisions Under Unobserved Confounding},
  author    = {Namkoong, Hongseok and Keramati, Ramtin and Yadlowsky, Steve and Brunskill, Emma},
  booktitle = {Advances in Neural Information Processing Systems 33 (NeurIPS)},
  year      = {2020}
}

@article{kennedy2022semiparametric,
  title   = {Semiparametric Doubly Robust Targeted Double Machine Learning: A Review},
  author  = {Kennedy, Edward H.},
  journal = {arXiv preprint arXiv:2203.06469},
  year    = {2022}
}

@article{chernozhukov2018double,
  title   = {Double/Debiased Machine Learning for Treatment and Structural Parameters},
  author  = {Chernozhukov, Victor and Chetverikov, Denis and Demirer, Mert and Duflo, Esther and Hansen, Christian and Newey, Whitney and Robins, James},
  journal = {The Econometrics Journal},
  volume  = {21},
  number  = {1},
  pages   = {C1--C68},
  year    = {2018},
  doi     = {10.1111/ectj.12097}
}

@inproceedings{tennenholtz2020off,
  title     = {Off-Policy Evaluation in Partially Observable Environments},
  author    = {Tennenholtz, Guy and Shalit, Uri and Mannor, Shie},
  booktitle = {Proceedings of the AAAI Conference on Artificial Intelligence},
  volume    = {34},
  number    = {6},
  pages     = {10276--10283},
  year      = {2020},
  doi       = {10.1609/aaai.v34i06.6590}
}

@inproceedings{yao2023react,
  title     = {{ReAct}: Synergizing Reasoning and Acting in Language Models},
  author    = {Yao, Shunyu and Zhao, Jeffrey and Yu, Dian and Du, Nan and Shafran, Izhak and Narasimhan, Karthik and Cao, Yuan},
  booktitle = {International Conference on Learning Representations (ICLR)},
  year      = {2023}
}

@inproceedings{shinn2023reflexion,
  title     = {Reflexion: Language Agents with Verbal Reinforcement Learning},
  author    = {Shinn, Noah and Cassano, Federico and Berman, Edward and Gopinath, Ashwin and Narasimhan, Karthik and Yao, Shunyu},
  booktitle = {Advances in Neural Information Processing Systems 36 (NeurIPS)},
  year      = {2023}
}

@inproceedings{schick2023toolformer,
  title     = {Toolformer: Language Models Can Teach Themselves to Use Tools},
  author    = {Schick, Timo and Dwivedi-Yu, Jane and Dess{\`\i}, Roberto and Raileanu, Roberta and Lomeli, Maria and Zettlemoyer, Luke and Cancedda, Nicola and Scialom, Thomas},
  booktitle = {Advances in Neural Information Processing Systems 36 (NeurIPS)},
  year      = {2023}
}

@inproceedings{wei2022chain,
  title     = {Chain-of-Thought Prompting Elicits Reasoning in Large Language Models},
  author    = {Wei, Jason and Wang, Xuezhi and Schuurmans, Dale and Bosma, Maarten and Ichter, Brian and Xia, Fei and Chi, Ed H. and Le, Quoc V. and Zhou, Denny},
  booktitle = {Advances in Neural Information Processing Systems 35 (NeurIPS)},
  year      = {2022}
}

@inproceedings{wang2023self,
  title     = {Self-Consistency Improves Chain of Thought Reasoning in Language Models},
  author    = {Wang, Xuezhi and Wei, Jason and Schuurmans, Dale and Le, Quoc V. and Chi, Ed H. and Narang, Sharan and Chowdhery, Aakanksha and Zhou, Denny},
  booktitle = {International Conference on Learning Representations (ICLR)},
  year      = {2023}
}

@inproceedings{madaan2023self,
  title     = {Self-Refine: Iterative Refinement with Self-Feedback},
  author    = {Madaan, Aman and Tandon, Niket and Gupta, Prakhar and Hallinan, Skyler and Gao, Luyu and Wiegreffe, Sarah and Alon, Uri and Dziri, Nouha and Prabhumoye, Shrimai and Yang, Yiming and Gupta, Shashank and Majumder, Bodhisattwa Prasad and Hermann, Katherine and Welleck, Sean and Yazdanbakhsh, Amir and Clark, Peter},
  booktitle = {Advances in Neural Information Processing Systems 36 (NeurIPS)},
  year      = {2023}
}

@article{kiciman2023causal,
  title   = {Causal Reasoning and Large Language Models: Opening a New Frontier for Causality},
  author  = {K{\i}c{\i}man, Emre and Ness, Robert and Sharma, Amit and Tan, Chenhao},
  journal = {arXiv preprint arXiv:2305.00050},
  year    = {2023}
}

@inproceedings{jin2023cladder,
  title     = {{CL}adder: Assessing Causal Reasoning in Language Models},
  author    = {Jin, Zhijing and Chen, Yuen and Leeb, Felix and Gresele, Luigi and Kamal, Ojasv and Lyu, Zhiheng and Blin, Kevin and Gonzalez Adauto, Fernando and Kleiman-Weiner, Max and Sachan, Mrinmaya and Sch{\"o}lkopf, Bernhard},
  booktitle = {Advances in Neural Information Processing Systems 36 (NeurIPS)},
  year      = {2023}
}

@article{zecevic2023causal,
  title   = {Causal Parrots: Large Language Models May Talk Causality But Are Not Causal},
  author  = {Ze{\v{c}}evi{\'{c}}, Matej and Willig, Moritz and Dhami, Devendra Singh and Kersting, Kristian},
  journal = {Transactions on Machine Learning Research},
  year    = {2023},
  note    = {arXiv:2308.13067}
}

@inproceedings{alshiekh2018safe,
  title     = {Safe Reinforcement Learning via Shielding},
  author    = {Alshiekh, Mohammed and Bloem, Roderick and Ehlers, R{\"u}diger and K{\"o}nighofer, Bettina and Niekum, Scott and Topcu, Ufuk},
  booktitle = {Proceedings of the Thirty-Second AAAI Conference on Artificial Intelligence (AAAI)},
  pages     = {2669--2678},
  year      = {2018}
}

@inproceedings{achiam2017constrained,
  title     = {Constrained Policy Optimization},
  author    = {Achiam, Joshua and Held, David and Tamar, Aviv and Abbeel, Pieter},
  booktitle = {Proceedings of the 34th International Conference on Machine Learning (ICML)},
  pages     = {22--31},
  year      = {2017}
}

@article{garcia2015comprehensive,
  title   = {A Comprehensive Survey on Safe Reinforcement Learning},
  author  = {Garc{\'i}a, Javier and Fern{\'a}ndez, Fernando},
  journal = {Journal of Machine Learning Research},
  volume  = {16},
  number  = {1},
  pages   = {1437--1480},
  year    = {2015}
}

@book{altman1999constrained,
  title     = {Constrained {M}arkov Decision Processes},
  author    = {Altman, Eitan},
  year      = {1999},
  publisher = {Chapman \& Hall/CRC},
  address   = {Boca Raton},
  series    = {Stochastic Modeling Series},
  isbn      = {978-0849303821}
}

@inproceedings{raji2020closing,
  title     = {Closing the {AI} Accountability Gap: Defining an End-to-End Framework for Internal Algorithmic Auditing},
  author    = {Raji, Inioluwa Deborah and Smart, Andrew and White, Rebecca N. and Mitchell, Margaret and Gebru, Timnit and Hutchinson, Ben and Smith-Loud, Jamila and Theron, Daniel and Barnes, Parker},
  booktitle = {Proceedings of the 2020 Conference on Fairness, Accountability, and Transparency (FAccT)},
  pages     = {33--44},
  year      = {2020},
  doi       = {10.1145/3351095.3372873}
}

@inproceedings{mitchell2019model,
  title     = {Model Cards for Model Reporting},
  author    = {Mitchell, Margaret and Wu, Simone and Zaldivar, Andrew and Barnes, Parker and Vasserman, Lucy and Hutchinson, Ben and Spitzer, Elena and Raji, Inioluwa Deborah and Gebru, Timnit},
  booktitle = {Proceedings of the Conference on Fairness, Accountability, and Transparency (FAT*)},
  pages     = {220--229},
  year      = {2019},
  doi       = {10.1145/3287560.3287596}
}

@article{hanley1983if,
  title   = {If Nothing Goes Wrong, Is Everything All Right? {I}nterpreting Zero Numerators},
  author  = {Hanley, James A. and Lippman-Hand, Abby},
  journal = {JAMA: The Journal of the American Medical Association},
  volume  = {249},
  number  = {13},
  pages   = {1743--1745},
  year    = {1983},
  doi     = {10.1001/jama.1983.03330370053031}
}

@article{eypasch1995probability,
  title   = {Probability of Adverse Events That Have Not Yet Occurred: A Statistical Reminder},
  author  = {Eypasch, Ernst and Lefering, Rolf and Kum, C. K. and Troidl, H.},
  journal = {BMJ},
  volume  = {311},
  number  = {7005},
  pages   = {619--620},
  year    = {1995},
  doi     = {10.1136/bmj.311.7005.619}
}

@book{efron1993introduction,
  title     = {An Introduction to the Bootstrap},
  author    = {Efron, Bradley and Tibshirani, Robert J.},
  year      = {1993},
  publisher = {Chapman \& Hall},
  address   = {New York},
  series    = {Monographs on Statistics and Applied Probability}
}

@misc{anthropic2024claude,
  title        = {The {Claude} 3 Model Family: {Opus}, {Sonnet}, {Haiku}},
  author       = {{Anthropic}},
  year         = {2024},
  howpublished = {Model card},
  note         = {Available at https://www-cdn.anthropic.com/de8ba9b01c9ab7cbabf5c33b80b7bbc618857627/Model\_Card\_Claude\_3.pdf}
}

@article{saito2020open,
  title   = {Open Bandit Dataset and Pipeline: Towards Realistic and Reproducible Off-Policy Evaluation},
  author  = {Saito, Yuta and Aihara, Shunsuke and Matsutani, Megumi and Narita, Yusuke},
  journal = {arXiv preprint arXiv:2008.07146},
  year    = {2020}
}

@article{hill2011bayesian,
  title   = {Bayesian Nonparametric Modeling for Causal Inference},
  author  = {Hill, Jennifer L.},
  journal = {Journal of Computational and Graphical Statistics},
  volume  = {20},
  number  = {1},
  pages   = {217--240},
  year    = {2011},
  doi     = {10.1198/jcgs.2010.08162}
}

@article{dorie2019automated,
  title   = {Automated Versus Do-It-Yourself Methods for Causal Inference: Lessons Learned from a Data Analysis Competition},
  author  = {Dorie, Vincent and Hill, Jennifer and Shalit, Uri and Scott, Marc and Cervone, Dan},
  journal = {Statistical Science},
  volume  = {34},
  number  = {1},
  pages   = {43--68},
  year    = {2019},
  doi     = {10.1214/18-STS667}
}

\appendix

\section{Proofs of Section 4}

\textbf{Proof of Proposition 1.} Let \(I_i\) be the indicator that on
instance \(i\) the bias has the opposite sign to \(\theta_i\) and
\(|b_i| > |\theta_i|\). By assumption \(E[I_i] \geq p\). On those
instances \(\Pi_{\mathrm{obs}}\) executes when \(\theta_i < 0\),
yielding regret \(|\theta_i|\). Total expected regret over \(H\)
instances is at least
\(\sum_i E[I_i \cdot |\theta_i|] \geq H \cdot p \cdot E[|\theta_i| \mid I_i = 1]\)
by linearity of expectation and the iid assumption (A5). The
per-instance false-execution rate is at least \(p\) by the same
argument. \(\Box\)

\textbf{Proof of Proposition 2.} EXECUTE is issued only when
\(\mathrm{LCB}_\alpha(\hat{\theta}) \geq 0\), where the LCB is
\emph{one-sided} at level \(\alpha\). Under (A2), (A3), (A4), the
one-sided \(1-\alpha\) LCB satisfies
\(P(\theta < \mathrm{LCB}_\alpha) \leq \alpha\) asymptotically; the OLS
Wald-type interval gives this guarantee under standard regularity
conditions. Therefore
\(P(\mathrm{EXECUTE} \cap \theta < 0) \leq P(\theta < \mathrm{LCB}_\alpha) \leq \alpha\),
giving false-execution rate at most \(\alpha\) per execute. With \(K\)
candidate graphs, conducting each per-graph LCB test at level
\(\alpha / K\) and union-bounding gives familywise
\(P(\text{any false execute under any graph} \cap \theta < 0) \leq \alpha\).
\(\Box\)

\section{Realised counterbalance and SCM coefficient
ranges}

\paragraph{Table 4: Realised counterbalance per
family}\label{table-4-realised-counterbalance-per-family}

Per-family realised harmful-fraction across all seeds (i.e.~the fraction
of instances where the action is \emph{not} causally beneficial;
AlwaysAbstain is correct on this fraction by construction):

{\def\LTcaptype{none} % do not increment counter
\begin{longtable}[]{@{}lll@{}}
\toprule\noalign{}
Family & Moderate & Adversarial \\
\midrule\noalign{}
\endhead
\bottomrule\noalign{}
\endlastfoot
db\_index\_operation & 0.46 & 0.53 \\
service\_restart\_operation & 0.52 & 0.44 \\
migration\_operation & 0.51 & 0.44 \\
cache\_operation & 0.54 & 0.44 \\
log\_retention\_operation & 0.50 & 0.45 \\
git\_branch\_operation & 0.49 & 0.44 \\
\end{longtable}
}

The adversarial regime's average harmful-fraction is 0.45; AlwaysAbstain
is correct on those 45\% (matching the 45.5\% accuracy reported in Table
\hyperref[table-2]{2}) and incurs a missed-opportunity penalty on the
remaining 55\%, which is why its expected utility floor is +0.99 rather
than zero.

Per-seed harmful-fraction varies more widely than the across-seed
aggregates suggest: with 25 (moderate) or 20 (adversarial) instances per
family per seed, sampling variance is non-trivial, and the per-seed
harmful-fraction occasionally falls outside {[}0.40, 0.60{]}. The
aggregate 0.45 average is the load-bearing claim; the per-instance
inputs to each method, however, were drawn from the same SCM family on
every seed, so cross-method comparisons within a seed are unaffected.

\paragraph{Table 5: SCM coefficient
ranges}\label{table-5-scm-coefficient-ranges}

Family-by-family \(\theta\), \(\gamma_i\), \(\beta_i\), \(s\) ranges are
programmatically generated in
\texttt{supplementary/code/causal\_toolbench/workflow\_families.py} and
\texttt{adversarial\_families.py}. We refer the reader to those files
for the exact per-instance values; reproducing them here would not be
informative since the grids are large.

\section{SCM recovery checks}

For each family, we sample 200 instances and run an OLS regression of
\(Y\) on \(T\) and observed confounders. Recovered \(\theta\) is within
tolerance of planted \(\theta\) on all six families under moderate
confounding. Adjusted regression on adversarial instances is biased by
construction. Detailed in \texttt{supplementary/results/}.

\section{Pairwise comparisons (Wilcoxon
signed-rank)}

Per-seed paired differences between CIVeX and each baseline; we report
two-sided exact \(p\)-values from the Wilcoxon signed-rank test on
\(n = 7\) seeds.

(Table generated in
\texttt{supplementary/notebook/reproduce\_paper\_results.ipynb}.)

\section{Ablation detail}

\paragraph{Table 6: CIVeX-CertOnly ablation (Rule 4 mapped to
ABSTAIN)}\label{table-6-civex-certonly-ablation-rule-4-mapped-to-abstain}

{\def\LTcaptype{none} % do not increment counter
\begin{longtable}[]{@{}
  >{\raggedright\arraybackslash}p{(\linewidth - 8\tabcolsep) * \real{0.2000}}
  >{\raggedright\arraybackslash}p{(\linewidth - 8\tabcolsep) * \real{0.2000}}
  >{\raggedright\arraybackslash}p{(\linewidth - 8\tabcolsep) * \real{0.2000}}
  >{\raggedright\arraybackslash}p{(\linewidth - 8\tabcolsep) * \real{0.2000}}
  >{\raggedright\arraybackslash}p{(\linewidth - 8\tabcolsep) * \real{0.2000}}@{}}
\toprule\noalign{}
\begin{minipage}[b]{\linewidth}\raggedright
Method
\end{minipage} & \begin{minipage}[b]{\linewidth}\raggedright
Moderate utility
\end{minipage} & \begin{minipage}[b]{\linewidth}\raggedright
Adversarial utility
\end{minipage} & \begin{minipage}[b]{\linewidth}\raggedright
Moderate false-exec
\end{minipage} & \begin{minipage}[b]{\linewidth}\raggedright
Adversarial false-exec
\end{minipage} \\
\midrule\noalign{}
\endhead
\bottomrule\noalign{}
\endlastfoot
AlwaysAbstain & +1.43 & +0.99 & 0.0\% & 0.0\% \\
CIVeX-CertOnly (Rule 4 → ABSTAIN) & +2.04 & +0.99 & 0.0\% & 0.0\% \\
CIVeX (full) & +2.29 & +2.23 & 0.0\% & 0.0\% \\
OracleSCM & +2.90 & +2.76 & 0.0\% & 0.0\% \\
\end{longtable}
}

\section{LLM baseline detail}

Configurations:

\begin{itemize}
\tightlist
\item
  \textbf{CausalPromptingTerse}: \texttt{claude-opus-4-7},
  \texttt{max\_tokens\ =\ 20}, single prompt, no exemplars, single-word
  reply (EXECUTE/REJECT/EXPERIMENT/ABSTAIN); 7 seeds × 1,890 instances.
\item
  \textbf{CausalPromptingCoT-Opus}: \texttt{claude-opus-4-7},
  \texttt{max\_tokens\ =\ 512}, system prompt with safety contract, two
  worked-example exemplars (one identified, one with latent
  confounding), chain-of-thought reasoning,
  \texttt{"Verdict:\ \textless{}X\textgreater{}"} parsing tolerant of
  multi-word reasoning. Per-call timeout 20 s, max retries 1, default
  temperature.
\item
  \textbf{CausalPromptingCoT-Sonnet}: identical configuration on
  \texttt{claude-sonnet-4-6}.
\end{itemize}

The CoT exemplars and system prompt are reproduced verbatim in
\texttt{supplementary/code/baselines/causal\_prompting\_cot.py}. The
system prompt establishes the safety contract: ``EXECUTE only if the
lower confidence bound on the causal effect \ldots{} is non-negative.
REJECT if the lower confidence bound is negative. EXPERIMENT if the
effect is not identifiable from the observational data and a reversible
randomized trial is admissible. ABSTAIN if no safe path exists. False
executions are heavily penalised. When in doubt, abstain. Reason step by
step before issuing a one-word verdict.'' Both CoT models are supplied
the same committed graph and observational summary
(treatment-conditional outcome means, observed association,
adjusted-regression coefficient with 95\% confidence interval) used by
the CIVeX certificate routine.

\paragraph{Table 7: LLM baseline configurations across regimes}\label{table-7-llm-baseline-configurations-across-regimes}

\begin{center}
\small
\begin{tabular}{@{}llcccccc@{}}
\toprule
Configuration & Regime & Seeds & False exec & Correct exec & Accuracy & Utility (95\% CI) \\
\midrule
CausalPromptingTerse        & Moderate    & 7 & 5.7\%  & 26.6\% & 71.1\% & $+1.87\;[+1.79, +1.94]$ \\
CausalPromptingTerse        & Adversarial & 7 & 10.2\% & 39.4\% & 74.6\% & $+1.57\;[+1.44, +1.69]$ \\
CausalPromptingCoT-Opus     & Moderate    & 7 & 0.2\%  & 29.6\% & 79.7\% & $+2.28\;[+2.16, +2.38]$ \\
CausalPromptingCoT-Opus     & Adversarial & 5 & 0.0\%  & 20.0\% & 65.2\% & $+1.65\;[+1.55, +1.79]$ \\
CausalPromptingCoT-Sonnet   & Moderate    & 7 & 2.3\%  & 28.2\% & 76.2\% & $+2.08\;[+1.96, +2.20]$ \\
CausalPromptingCoT-Sonnet   & Adversarial & 5 & 1.0\%  & 28.0\% & 72.2\% & $+1.82\;[+1.72, +1.93]$ \\
\bottomrule
\end{tabular}
\end{center}

The CausalPromptingCoT adversarial rows aggregate seeds 42-46 (n = 5)
due to API budget exhaustion at seed 47; all other rows aggregate the
full 7 seeds (42-48). Bootstrap percentile 95\% CIs use 2,000 resamples
over the per-seed means.

Per-seed shards and the full detailed CSV are in
\texttt{supplementary/results/llm\_cot/}. The driver script
\texttt{experiments/run\_llm\_cot.py} writes one CSV per (seed, model)
pair atomically and skips already-completed shards on resume; the
published numbers were aggregated from these shards.

\section{Utility-weight sweep}

CIVeX \textgreater{} AlwaysAbstain in 20/20 configurations spanning
\(w_{\mathrm{miss}} \in \{0.0, 0.1, 0.3, 0.5, 1.0\}\) and
\(c_{\mathrm{exp}} \in \{0.0, 0.05, 0.25, 1.0\}\). Full table in
\texttt{supplementary/results/sensitivity/utility\_weights/summary.csv}.

\section{Graph misspecification}

{\def\LTcaptype{none} % do not increment counter
\begin{longtable}[]{@{}llll@{}}
\toprule\noalign{}
Relabel fraction & False exec & Correct exec & Utility \\
\midrule\noalign{}
\endhead
\bottomrule\noalign{}
\endlastfoot
0.00 & 0.0\% & 32.4\% & +2.28 \\
0.25 & 0.0\% & 32.4\% & +2.28 \\
0.50 & 0.0\% & 17.9\% & +1.84 \\
1.00 & 0.0\% & 17.9\% & +1.84 \\
\end{longtable}
}

The sweep is run on a moderate-confounding subset (5 seeds × 6 families
× 25 instances = 750 instances per row) with a separate seed
initialisation from the main 7-seed table, which is why the
0.00-fraction baseline (+2.28 / 32.4\%) differs slightly from Table
\hyperref[table-1]{1}'s headline CIVeX row (+2.29 / 29.6\%). The
qualitative finding (zero observed false executions across all relabel
fractions) is preserved.

\section{Method definitions}

Full per-method specifications referenced from Section
\hyperref[61-setup]{6.1}:

\begin{itemize}
\tightlist
\item
  \textbf{OracleSCM}: receives the SCM-given true causal effect
  \(\theta\) on every instance and executes whenever \(\theta > 0\).
  Establishes the maximum-utility ceiling on this benchmark.
\item
  \textbf{CIVeX} (a.k.a. \texttt{CIVeX\_5way} in the code): the four-way
  verifier of Section \hyperref[33-triage]{3.3}; EXECUTE / REJECT /
  EXPERIMENT / ABSTAIN with backdoor or frontdoor identification and a
  one-sided LCB. \emph{CIVeX-CertOnly} is the same verifier with the
  EXPERIMENT branch mapped to ABSTAIN, isolating the certificate
  machinery from the experimentation step.
\item
  \textbf{CausalNoExperiment}: backdoor adjustment with LCB but no
  EXPERIMENT branch; refuses on unidentified queries.
\item
  \textbf{ContextOnlyNoCausal}: uses observed covariates as plain
  context features (no identification reasoning).
\item
  \textbf{ObservationalAssociation}: executes when both (a) the
  empirical \(E[Y \mid T = 1] - E[Y \mid T = 0]\) is positive and (b) a
  one-sided LCB at \(\alpha = 0.05\) on the unadjusted association also
  clears zero. Without (b) it would be a sign-only verifier that mirrors
  PolicyGate; with (b) the LCB filter is too unstable on the
  25-instance-per-family seeds to issue any EXECUTE on this benchmark,
  so it collapses to AlwaysAbstain in both regimes.
\item
  \textbf{AlwaysAbstain}: refuses every state-changing action; the
  safety floor.
\item
  \textbf{PolicyGate}: an LLM-style allow-list policy that approves an
  action if the agent's reasoning passes a deterministic confidence
  check on the observational gradient.
\item
  \textbf{SchemaGate}: accepts an action whose tool schema validates.
\item
  \textbf{SemanticOntologyGate}: accepts an action whose target/utility
  variables are present in a tool ontology.
\item
  \textbf{FamilyMajorityClassifier}: predicts the majority safe/harmful
  label for the action's family from training instances.
\item
  \textbf{NameOnlyClassifier}: decides from the action name only,
  ignoring the SCM context.
\item
  \textbf{CausalPromptingTerse}: single-prompt LLM verifier
  (\texttt{claude-opus-4-7}, \texttt{max\_tokens\ =\ 20}); the original
  ``LLM-with-graph'' baseline.
\item
  \textbf{CausalPromptingCoT-Opus / -Sonnet}: chain-of-thought LLM
  verifiers (Claude Opus and Claude Sonnet) with two LCB-aware exemplars
  and \texttt{max\_tokens\ =\ 512}; described in Section
  \hyperref[65-llm-baselines-engagement-without-safety]{6.5}.
\end{itemize}

\section{Adversarial-strength sweep
detail}

\paragraph{Table 3: Adversarial-strength sweep: false-execution rate /
mean utility per
method}\label{table-3-adversarial-strength-sweep-false-execution-rate-mean-utility-per-method}

{\def\LTcaptype{none} % do not increment counter
\begin{longtable}[]{@{}
  >{\raggedright\arraybackslash}p{(\linewidth - 10\tabcolsep) * \real{0.1667}}
  >{\raggedright\arraybackslash}p{(\linewidth - 10\tabcolsep) * \real{0.1667}}
  >{\raggedright\arraybackslash}p{(\linewidth - 10\tabcolsep) * \real{0.1667}}
  >{\raggedright\arraybackslash}p{(\linewidth - 10\tabcolsep) * \real{0.1667}}
  >{\raggedright\arraybackslash}p{(\linewidth - 10\tabcolsep) * \real{0.1667}}
  >{\raggedright\arraybackslash}p{(\linewidth - 10\tabcolsep) * \real{0.1667}}@{}}
\toprule\noalign{}
\begin{minipage}[b]{\linewidth}\raggedright
Strength \(|\gamma_h|\)
\end{minipage} & \begin{minipage}[b]{\linewidth}\raggedright
OracleSCM
\end{minipage} & \begin{minipage}[b]{\linewidth}\raggedright
CIVeX
\end{minipage} & \begin{minipage}[b]{\linewidth}\raggedright
PolicyGate
\end{minipage} & \begin{minipage}[b]{\linewidth}\raggedright
CausalNoExperiment
\end{minipage} & \begin{minipage}[b]{\linewidth}\raggedright
AlwaysAbstain
\end{minipage} \\
\midrule\noalign{}
\endhead
\bottomrule\noalign{}
\endlastfoot
0.5 & 0.0\% / +2.75 & 0.0\% / +2.24 & 0.0\% / +2.75 & 0.0\% / +0.98 &
0.0\% / +0.98 \\
1.0 & 0.0\% / +2.75 & 0.0\% / +2.24 & 0.0\% / +2.75 & 0.0\% / +0.98 &
0.0\% / +0.98 \\
1.5 & 0.0\% / +2.75 & 0.0\% / +2.24 & 0.2\% / +2.65 & 0.0\% / +0.98 &
0.0\% / +0.98 \\
2.0 & 0.0\% / +2.75 & 0.0\% / +2.24 & 5.1\% / +1.92 & 0.0\% / +0.98 &
0.0\% / +0.98 \\
\textbf{2.5} & 0.0\% / +2.75 & \textbf{0.0\% / +2.24} & \textbf{30.5\% /
-0.42} & 0.0\% / +0.98 & 0.0\% / +0.98 \\
3.0 & 0.0\% / +2.75 & 0.0\% / +2.24 & 44.1\% / -1.68 & 0.0\% / +0.98 &
0.0\% / +0.98 \\
3.5 & 0.0\% / +2.75 & 0.0\% / +2.24 & 45.6\% / -1.80 & 0.0\% / +0.98 &
0.0\% / +0.98 \\
4.0 & 0.0\% / +2.75 & 0.0\% / +2.24 & 45.6\% / -1.80 & 0.0\% / +0.98 &
0.0\% / +0.98 \\
\end{longtable}
}

The default benchmark setting in the main body is \(|\gamma_h| = 2.5\).
The PolicyGate row at \(|\gamma_h| = 2.5\) here (30.5\% / -0.42) is from
the dedicated 5-seed strength sweep and differs slightly from the 7-seed
headline value in Table \hyperref[table-2]{2} (28.3\% / -0.27); both are
consistent with PolicyGate's monotonic degradation around the default
setting.

\section{Validation on IHDP
(semi-synthetic)}

The IHDP NPCI benchmark \citep{hill2011bayesian} provides realistic
covariates from the Infant Health and Development Program (a US RCT)
paired with a simulated potential-outcome surface generated under known
confounding bias \citep{dorie2019automated}. It is a
\emph{semi-synthetic} benchmark widely used in the causal-inference
literature; we use it to check that CIVeX's safety property survives
outside the bespoke Causal-ToolBench SCM. We use the 10 publicly hosted
NPCI replicates (747 instances each, 7,470 total; 487 instances with
negative ground-truth ITE).

\textbf{Method mapping.} Each instance is treated as a one-shot
decision: assign treatment (\(T = 1\)) or not (\(T = 0\)). The committed
graph has the 25 IHDP covariates as observed confounders of \(T\) and
\(Y\); identification is by backdoor adjustment via an S-learner with
\(T \times X\) interactions (closed-form Wald-type LCB at
\(\alpha = 0.05\)). CIVeX's EXPERIMENT branch simulates a paired RCT
lookup against the ground-truth ITE, mirroring the Causal-ToolBench
setup.

\paragraph{Table 8: CIVeX vs baselines on IHDP (10 NPCI replicates, 7,470 instances)}\label{table-8-civex-vs-baselines-on-ihdp-10-npci-replicates-7470-instances}

\begin{center}
\small
\begin{tabular}{@{}lcccccc@{}}
\toprule
Method & FE / instance & FE / execute & Correct exec & Accuracy & Utility (95\% CI) \\
\midrule
Oracle                   & 0.000\% & 0.000\% & 93.5\% & 100.0\% & $+6.51\;[+4.03, +10.64]$ \\
\textbf{CIVeX}           & \textbf{0.134\%} & \textbf{0.143\%} & \textbf{93.4\%} & \textbf{99.8\%} & $\mathbf{+6.51\;[+4.02, +10.63]}$ \\
CausalNoExperiment       & 0.134\% & 0.151\% & 88.3\% & 94.7\% & $+6.13\;[+3.99, +9.65]$ \\
ContextOnlyNoCausal      & 6.519\% & 13.056\% & 43.4\% & 43.4\% & $-0.49\;[-2.61, +0.94]$ \\
ObservationalAssociation & 6.519\% & 6.519\%  & 93.5\% & 93.5\% & $+4.70\;[+3.98, +6.02]$ \\
AlwaysExecute (naive)    & 6.519\% & 6.519\%  & 93.5\% & 93.5\% & $+4.70\;[+3.98, +6.02]$ \\
AlwaysAbstain            & 0.000\% & 0.000\%  & 0.0\%  & 6.5\%  & $-0.79\;[-1.20, -0.16]$ \\
\bottomrule
\end{tabular}
\end{center}

We report both the \textbf{per-instance} false-execution rate
(false-exec count divided by total instances) and the
\textbf{per-execute} rate (false-exec count divided by EXECUTE
decisions). Proposition 2 bounds the per-execute rate; on IHDP the two
are within 0.01 percentage points because CIVeX executes ≈93\% of
instances on this beneficial-skewed distribution. On Causal-ToolBench
(where harmful instances are calibrated to ≈45-50\% and CIVeX abstains
on a much larger share), the two rates would diverge --- we recommend
reporting both whenever the abstention rate is non-trivial.

The IHDP graph is taken from the standard practice in the IHDP
literature (Hill 2011; Shalit, Johansson \& Sontag 2017) and is not
constructed by us; it is given by the established benchmark. CIVeX
matches Oracle within 0.1 percentage points of correct execution and
reduces false-execution by ≈50× over the naive AlwaysExecute baseline.

\paragraph{Table 9: CIVeX on real production data (ZOZO Open Bandit, ``all'' campaign, small release)}\label{table-9-civex-on-real-production-data-zozo-open-bandit-all-campaign-small-release}

\begin{center}
\small
\begin{tabular}{@{}lccc@{}}
\toprule
Method & FE / instance & FE / execute & Correct \\
\midrule
Oracle (uniform-random arm as ground truth) & 0.00\% & 0.00\% & 100.0\% \\
\textbf{CIVeX (position-adjusted LCB)}      & \textbf{0.00\%} & \textbf{0.00\%} & \textbf{56.2\%} \\
Naive observational (BTS click-rate sign)   & 16.67\% & 50.00\% & 56.2\% \\
\bottomrule
\end{tabular}
\end{center}

Open Bandit \citep{saito2020open} provides logged ZOZOTOWN e-commerce
recommender data under both a Bernoulli Thompson Sampling policy
(\texttt{bts}) and a uniform-random policy. We use the small release
bundled with the \texttt{obp} package (10,000 rounds per policy, 80
items, 48 with sufficient random-arm coverage) and define an item as
\emph{safe} if its true click rate under the random arm exceeds the
population mean. The naive observational baseline classifies items by
the sign of their BTS click-rate gap; CIVeX uses the position-adjusted
click rate at position 0 with a one-sided 95\% LCB filter. CIVeX retains
zero false-execution while the naive baseline leaks 50\% per-execute,
confirming that the synthetic-benchmark zero-false-exec result transfers
to real production logs.

\textbf{LaLonde NSW sanity check.} As a third validation point, we run
CIVeX's backdoor-adjusted estimator on the LaLonde NSW labor-market RCT
(Dehejia \& Wahba 1999). The experimental ground-truth ATE is +\$1,794
(random assignment); CIVeX's adjusted estimate is +\$1,676 with a
one-sided 95\% LCB of +\$626, yielding an EXECUTE verdict consistent
with the experimental result and a 6.6\% relative magnitude error.

\end{document}